\documentclass[10pt,twocolumn,letterpaper]{article}

\usepackage{titling}
\usepackage{cvpr}
\usepackage{times}
\usepackage{epsfig}
\usepackage{graphicx}
\usepackage{amsmath}
\usepackage{amssymb}
\usepackage{algorithm}
\usepackage{algorithmic}
\usepackage{subcaption}
\usepackage{multirow}
\usepackage{color}
\usepackage{comment}
\newtheorem{prop}{Proposition}

\newtheorem{lem}{Lemma}
\newtheorem{define}{Definition}

\usepackage[breaklinks=true,bookmarks=false]{hyperref}

\cvprfinalcopy 

\def\cvprPaperID{2944} 

\makeatletter
\newcommand{\specificthanks}[1]{\@fnsymbol{#1}}
\makeatother
\ifcvprfinal\fi
\begin{document}

\title{Iterative Projection and Matching: Finding Structure-preserving Representatives and Its Application to Computer Vision}

\author{Mohsen Joneidi\thanks{indicates shared first authorship.}
 , Alireza Zaeemzadeh\textsuperscript{\specificthanks{1}}, Nazanin Rahnavard, and Mubarak Shah\\
University of Central Florida\\
{\tt\small \{joneidi, zaeemzadeh, nazanin\}@eecs.ucf.edu, shah@crcv.ucf.edu}
}

\maketitle

\begin{abstract}
The goal of data selection is to capture the most structural information from a set of data. 
This paper presents a fast and accurate data selection method, in which the selected samples are optimized to span the subspace of all data. We propose a new selection algorithm, referred to as iterative projection and matching (IPM), with linear complexity w.r.t. the number of data, and without any parameter to be tuned. In our algorithm,  at each iteration, the maximum information from the structure of the data is captured by one selected sample, and the captured information is neglected in the next iterations by projection on the null-space of previously selected samples.
The computational efficiency and the selection accuracy of our proposed algorithm outperform those of the conventional methods. Furthermore, the superiority of the proposed algorithm is shown on active learning for video action recognition dataset on UCF-101; learning using representatives on ImageNet; training a generative adversarial network (GAN) to generate multi-view images from a single-view input on CMU Multi-PIE dataset; and video summarization on UTE Egocentric dataset.
\end{abstract}

\section{Introduction}
\label{sec:intro}

Thanks to recent advances in computing, deep learning based systems, which employ very large numbers of inputs, have been developed in the last decade. However, processing/labeling/communication of a large number of input data has remained challenging. Therefore, novel machine learning algorithms that  make the best use of a significantly less amount of data are of great interest. For example, active learning (AL)~\cite{lewis1994sequential} aims at addressing this problem by iteratively training a model using a small number of labeled data, testing the whole data on the trained model, and then querying the labels of some selected data, which then are used for training a new model.  In this context, preserving the underlying structure of data by a succinct format is an essential concern. 

Data selection task is not trivial and possibly implies addressing an NP-hard problem (i.e., there are $M\choose K$ possibilities of choosing $K$ distinct sample out of $M$ available ones). This means that an optimal solution cannot be efficiently computed when the number of available data becomes excessively large. A convex relaxation of the original NP-hard problem has been suggested in terms of the {D-optimal} and {A-optimal} solutions \cite{Boyd:2004:CO:993483,Joshi2009Sensor_selection}.  In addition to convex relaxation, a sub-modular cost function as the criterion of selection, allows us to employ much faster greedy optimization methods for selection \cite{sens_sel_greedy5717225}. The stochastic implementation of D-optimal solution is referred to \emph{volume sampling} (VS), which is a fast and well-studied method. VS selects each subset of data, which are organized in the rows of a matrix, with probability proportional to the determinant (volume) of the reduced matrix. Moreover, QR decomposition with column pivoting (QRCP) and convex hull-based selection methods have been suggested for optimal data selection \cite{duersch2017randomized,ding2018fast}. All the mentioned methods aim to select the most diverse subset of data in an optimal sense. However, these methods do not guarantee that the un-selected samples are well-covered by the selected ones. Further,  outliers are selected with a high probability using such algorithms due to their diversity. Authors in \cite{joneidi2018Eoptimal} address this problem via a two-phase algorithm. There are some other efforts for outlier rejection in the selection procedure \cite{rahmani2017robust,wang2017representative}. However, the outlier and inlier data are not well-defined and these methods are not consistent with general data. 

 There is another more effective approach for subset selection, which chooses data such that the selected samples are able to approximate the rest of data accurately. This selection problem is formulated using a convex optimization problem and referred as sparse modeling representative selection (SMRS) algorithm \cite{Elhamifar2012SeeObjects}. The same goal is pursued by dissimilarity-based sparse subset selection (DS3), which is based on simultaneous sparse recovery for finding data representatives \cite{Elhamifar2016DissimilaritySelection}. Representative approaches, such as SMRS and DS3, provide more suitable subset rather than selecting some diverse samples. However, their computational burden is not tractable for large datasets. Moreover, SMRS and DS3 algorithms utilize some parameters in their implementation, which makes their fine tuning  difficult.

In order to address above issues, we propose a novel representative-based selection method, referred to as Iterative Projection and Matching (IPM). In our algorithm,  at each iteration the maximum information from structure of the data is captured by one selected sample, and the captured information is neglected in the next iterations   by projection on the null-space of previously selected samples. 
In summary, this paper makes the following contributions:
 \begin{itemize}
     \item The complexity of IPM is linear w.r.t. number of original data. Hence, IPM is tractable for larger datasets.
     \item IPM has no parameters for fine tuning, unlike some existing methods \cite{Elhamifar2016DissimilaritySelection,Elhamifar2012SeeObjects}. This makes IPM dataset- and problem-independent.
     \item Robustness of the proposed solution is investigated theoretically. 
     \item The superiority of the proposed algorithm is shown in different computer vision applications. 
 \end{itemize} 

\section{Problem Statement and Related Work}
\label{sec:pr}
Let $\boldsymbol{a}_1, \boldsymbol{a}_2,\ldots, \boldsymbol{a}_M\in \mathbb{R}^N$ be $M$ given data points of dimension $N$. We define an $M\times N$ matrix, $\boldsymbol{A}$, such that $\boldsymbol{a}_m^T$ is the $m^{th}$ row of A, for $m=1,2,\ldots, M$. The goal is to reduce this matrix into a $K\times N$ matrix, $\boldsymbol{A}_R$, based on an optimality metric. In this section, we introduce some related work on matrix subset selection and data selection. 

\subsection{ Selection Based on Diversity}
Consider a large system of equations $\boldsymbol{y}=\boldsymbol{Aw}$, which can be interpreted as a simple linear classifier in which $\boldsymbol{y}$ is the vector of labels, $\boldsymbol{A}$ represents the training data and $\boldsymbol{w}$ is the classifier weights. An optimal sense for data selection is to reduce this system of equations to a smaller system, $\boldsymbol{y}_R=\boldsymbol{A}_R\hat{\boldsymbol{w}}$, such that the reduced  subsystem estimates the same classifier as the original system, i.e., the estimation error of $\hat{\boldsymbol{w}}$ is minimized \cite{Derezinski2018SubsamplingSampling}.  
A typical selection objective is to minimize $\mathbb{E}_\nu\{\|\boldsymbol{w}-\hat{\boldsymbol{w}} \|_2\}$, where $\mathbb{E}_\nu$ is expectation w.r.t. noise distribution of $\boldsymbol{w}-\hat{\boldsymbol{w}}$. This criterion is referred as \emph{A-optimal} design in the literature of  optimization. It is an NP hard problem, which can be solved via convex relaxation with computational complexity of $O(M^3)$ \cite{Joshi2009Sensor_selection}.

However, there are other criteria which have interesting properties. For example \emph{D-optimal} design optimizes the determinant of a reduced  matrix \cite{Joshi2009Sensor_selection}. There are several other efforts in this area \cite{deshpande2010efficient,deshpande2006matrix,li2017polynomial,farahat2015greedy}. Inspired by D-optimal design, volume sampling (VS), which has received lots of attention, considers a selection probability for each subset of data, which is proportional to the determinant (volume) of the reduced matrix \cite{li2017polynomial,nikolov2018proportional,Derezinski2018SubsamplingSampling}. 
VS theory expresses that if \small{$\mathbb{T}\subset\{1,2, \ldots, M\}$} \normalsize{ is any subset with cardinality $K$, chosen with probability proportional to $\text{det}(\boldsymbol{A}_{\mathbb{T}}\boldsymbol{A}_{\mathbb{T}}^T)$, then\footnote{$\boldsymbol{A}_{\mathbb{T}}$ is the selected rows of $\boldsymbol{A}$  indexed by set $\mathbb{T}$.},}
\small{
\begin{equation}
\label{eq:VS}
\mathbb{E}\{ \|\boldsymbol{A}-\pi_{\mathbb{T}}(\boldsymbol{A})\|_F^2\}\le (K+1) \|\boldsymbol{A}-\boldsymbol{A}_K\|_F^2,
\end{equation}
}
\normalsize{where, $\pi_{\mathbb{T}}({\boldsymbol{A}})$ is a matrix representing projection of rows of $\boldsymbol{A}$ on to the span of selected rows indexed by $\mathbb{T}$. $\mathbb{E}$ indicates expectation operator w.r.t. all the combinatorial selection of $K$ rows of $\boldsymbol{A}$ out of $M$. $\boldsymbol{A}_K$ is the best rank-$K$ approximation of $\boldsymbol{A}$ that can be obtained by singular value decomposition and $\|.\|_F^2$ is the Frobenius norm. VS is not a deterministic selection algorithm, as it gives a probability of selection for any subset of samples, and for which only a loose upper bound for the expectation of projection error is guaranteed. In contrast, in this paper a deterministic algorithm is proposed based on direct minimization of projection error using a new optimization mechanism.
}

\subsection{Representative Selection}
A method for sampling from a set of data is proposed by Elhamifar et. al. based on sparse modeling representative selection (SMRS) \cite{Elhamifar2012SeeObjects}. Their proposed cost function for data selection is the error of projecting all the data onto the subspace spanned by the selected data. Mathematically, the optimization problem in~\cite{Elhamifar2012SeeObjects} can be written as,
\small{
\begin{equation}
\label{eq:orig_sel}
\underset{|\mathbb{T}|=K}{\text{argmin}}\|\boldsymbol{A}-\pi_{\mathbb{T}}(\boldsymbol{A})\|_F^2.
\end{equation}}
\normalsize{
 This  is an NP-hard problem. Their main contribution is solving this problem via convex relaxation.} 
However, there is no guarantee that convex relaxation provides the best approximation for an NP-hard problem. In this paper, we propose a new fast algorithm for solving Problem (\ref{eq:orig_sel}).


Dissimilarity-based Sparse Subset Selection (DS3) algorithm selects a subset of data based on pairwise distance of all data to some target points \cite{Elhamifar2016DissimilaritySelection}. DS3 considers a source dataset and its goal is to encode the target data according to pairwise dissimilarity between each sample of source and target datasets. This algorithm can be interpreted as the non-linear implementation of SMRS algorithm \cite{Elhamifar2016DissimilaritySelection}. 

\section{Iterative Projection and Matching (IPM)}
\label{sec:ipm}
In this section, an iterative and computationally efficient algorithm is proposed for approximating the solution to the NP-hard selection problem (\ref{eq:orig_sel}). The proposed algorithm iteratively finds the best direction on the unit sphere\footnote{In unit sphere, every point corresponds to a unique direction.}, and then from the available samples in dataset selects the sample with the smallest angle to the found direction. 

 Projection of all the data on to the subspace spanned by the $K$ rows of $\boldsymbol{A}$, indexed by $\mathbb{T}$, i.e., $\pi_{\mathbb{T}}(\boldsymbol{A})$, can be expressed by a rank-$K$ factorization, $\boldsymbol{U}\boldsymbol{V}^T$, where $\boldsymbol{U}\in \mathbb{R}^{M\times K}$, $\boldsymbol{V}^T\in \mathbb{R}^{K\times N}$, and $\boldsymbol{V}^T$ includes the $K$ rows of $\boldsymbol{A}$, indexed by $\mathbb{T}$, and normalized to have unit length. Therefore, optimization problem (\ref{eq:orig_sel}) can be restated as 
\begin{equation}
\small 
\label{eq:rewrite}
\underset{\boldsymbol{U},\boldsymbol{V}}{\text{argmin}}\;\|\boldsymbol{A}-\boldsymbol{UV}^T\|_F^2 \;\small{\text{s.t.}}\; \boldsymbol{v}_k\in \mathbb{A},
\end{equation}
where, $\mathbb{A}=\{\Tilde{\boldsymbol{a}_1},\Tilde{\boldsymbol{a}_2},\ldots, \Tilde{\boldsymbol{a}}_M\}$, $\Tilde{\boldsymbol{a}}_m=\boldsymbol{a}_m/\|\boldsymbol{a}_m\|_2$, and $\boldsymbol{v}_k$ is the $k^{\text{th}}$ column of $\boldsymbol{V}$. It should be noted that $\boldsymbol{V}^T$ is restricted to be a collection of $K$ normalized rows of $\boldsymbol{A}$, while there is no constraint on $\boldsymbol{U}$. Assume we are to select  one sample at a time, which is the best representation of all data. Since Problem (\ref{eq:rewrite}) involves a combinatorial search and is not easy to tackle, let us modify (\ref{eq:rewrite}) into two consecutive problems. The first sub-problem relaxes the constraint $\boldsymbol{v}_k \in \mathbb{A}$  in (\ref{eq:rewrite}) to a moderate constraint $\|\boldsymbol{v}\|=1$, and the second sub-problem reimposes the underlying constraint. These sub-problems are formulated as  

\begin{subequations}
\small
\begin{align}
\label{eq:our1}
(\boldsymbol{u},\boldsymbol{v})=&\underset{\boldsymbol{u},\boldsymbol{v}}{\text{argmin}}\;\|\boldsymbol{A}-\boldsymbol{uv}^T\|_F^2 \;\small{\text{s.t.}}\; \|\boldsymbol{v}\|=1,\\
\label{eq:our2}
m^{(1)}=&\underset{m}{\text{argmax}}\;| \boldsymbol{v}^T \Tilde{\boldsymbol{a}}_m|.
\end{align}
\label{eq:IPM_main}
\end{subequations}

 Here $m^{(1)}$ is the index of the first selected data point and $\boldsymbol{a}_{\boldsymbol{m}^{(1)}}$ is the selected sample. Subproblem (\ref{eq:our1}) is equivalent to finding the first right singular vector of $\boldsymbol{A}$. The constraint $\|\boldsymbol{v}\|=1$ keeps $\boldsymbol{v}$ on the unit sphere to remove scale ambiguity between $\boldsymbol{u}$ and $\boldsymbol{v}$. Moreover, the unit sphere is a superset for $\mathbb{A}$ and keeps the modified problem close to the recast problem (\ref{eq:rewrite}). After solving for $\boldsymbol{v}$ (which is not necessarily one of our data points), we find the data point that matches $\boldsymbol{v}$ the most (has the least angle with $\boldsymbol{v}$) in (\ref{eq:our2}). 
 
After selecting the first data point ($\boldsymbol{a}_{m^{(1)}}$), we project all data points onto the null space of the selected sample. This forms a new matrix $\boldsymbol{A}(\boldsymbol{I}-\Tilde{{\boldsymbol{a}}}_{m^{(1)}}\Tilde{{\boldsymbol{a}}}_{m^{(1)}}^T)$, where $\boldsymbol{I}$ is an identity matrix. We solve (\ref{eq:IPM_main}) with this new matrix to find the second data point. This process will continue until we select $K$ data points. It should be noted that the null space of selected sample(s) indicates a subspace that the selected sample(s) cannot span. Therefore, the next selected data is obtained by only searching in this null space.  
 
  Algorithm \ref{alg:IPM} shows the steps of the proposed iterative projection and matching (IPM) algorithm, in which $m^{(k)}$ denotes the index of the selected data at the $k^{th}$ iteration. IPM is a low-complexity algorithm with no parameters to be tuned. These features in addition to its superior performance (as will be shown in many scenarios in Section~\ref{sec:experiments}) make IPM  very desirable for a wide range of applications.  
  Time complexity order of computing the first singular component of an $M\times N$ matrix is $O(MN)$ \cite{Comon1990TrackingProcessing}. As the proposed algorithm only needs the first singular component for each selection, its time complexity  is $O(KNM)$, which is much faster than convex relaxation-based algorithms with complexity $O(M^3)$ \cite{Joshi2009Sensor_selection}.  Moreover, IPM performs faster than K-medoids algorithm,  whose complexity is of order $O(KN(M-K)^2)$ \cite{Vijaya2004Leaders--Subleaders:Sets}.

\begin{algorithm}
\caption{ \small Iterative Projection and Matching Algorithm}\label{alg:IPM}
\algsetup{
linenosize=\small,
linenodelimiter=:
}
\begin{algorithmic}[1]
\REQUIRE $\boldsymbol{A}$ and $K$\\
\hspace{-3mm} \textbf{Output}:$\;\boldsymbol{A}_{\mathbb{T}}$\\
\STATE \textbf{Initialization:}\\ $\boldsymbol{A}^{(1)} \longleftarrow \boldsymbol{A}$ \\
$\mathbb{T}=\{\}$
\\ $ \text{for}\; k=1, \cdots ,K $
\small{\STATE $\quad$ $v\xleftarrow[]{}$ first right singular-vector of $\boldsymbol{A}^{(k)}$ by solving  (\ref{eq:our1})
\STATE $\quad$ $m^{(k)}\xleftarrow[]{}$ index of the most correlated data with $\boldsymbol{v}$  (\ref{eq:our2})
\STATE $\quad$ $\mathbb{T}\xleftarrow[]{}\mathbb{T}\cup m^{(k)}$ \vspace{+1mm}
\STATE$\quad$ $\boldsymbol{A}^{(k+1)}\xleftarrow[]{}$ $\boldsymbol{A}^{(k)}(\boldsymbol{I}-\Tilde{{\boldsymbol{a}}}_{m^{(k)}}\Tilde{{\boldsymbol{a}}}_{m^{(k)}}^T)$ \small{(null space projection)} 
}
\\ end 
\end{algorithmic}
\end{algorithm}
\begin{figure}
\begin{subfigure}[b]{0.48\columnwidth}
   \includegraphics[width=1\columnwidth]{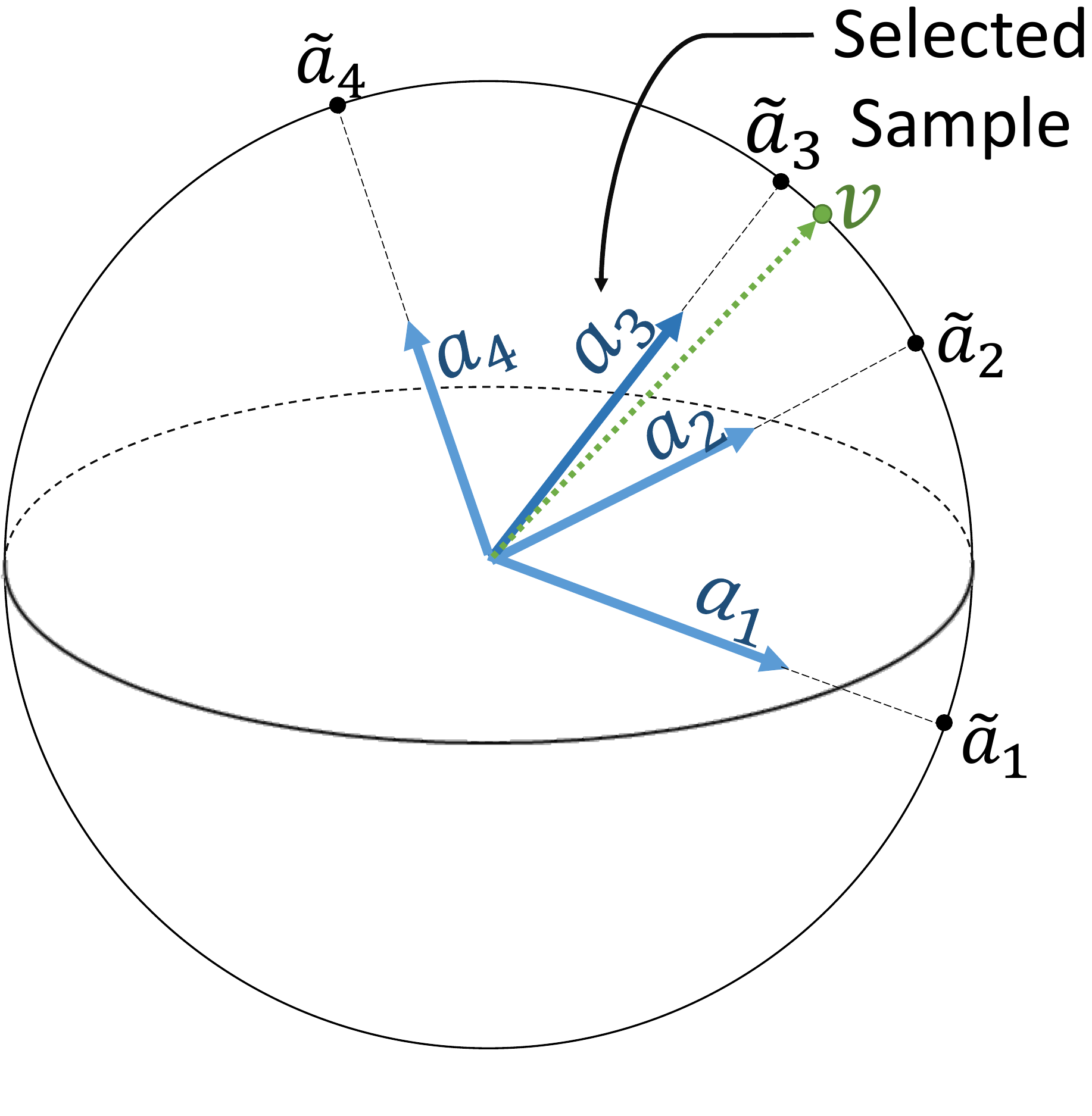}
\end{subfigure}
\begin{subfigure}[b]{0.48\columnwidth}
    \centering     
   \includegraphics[width=1\columnwidth]{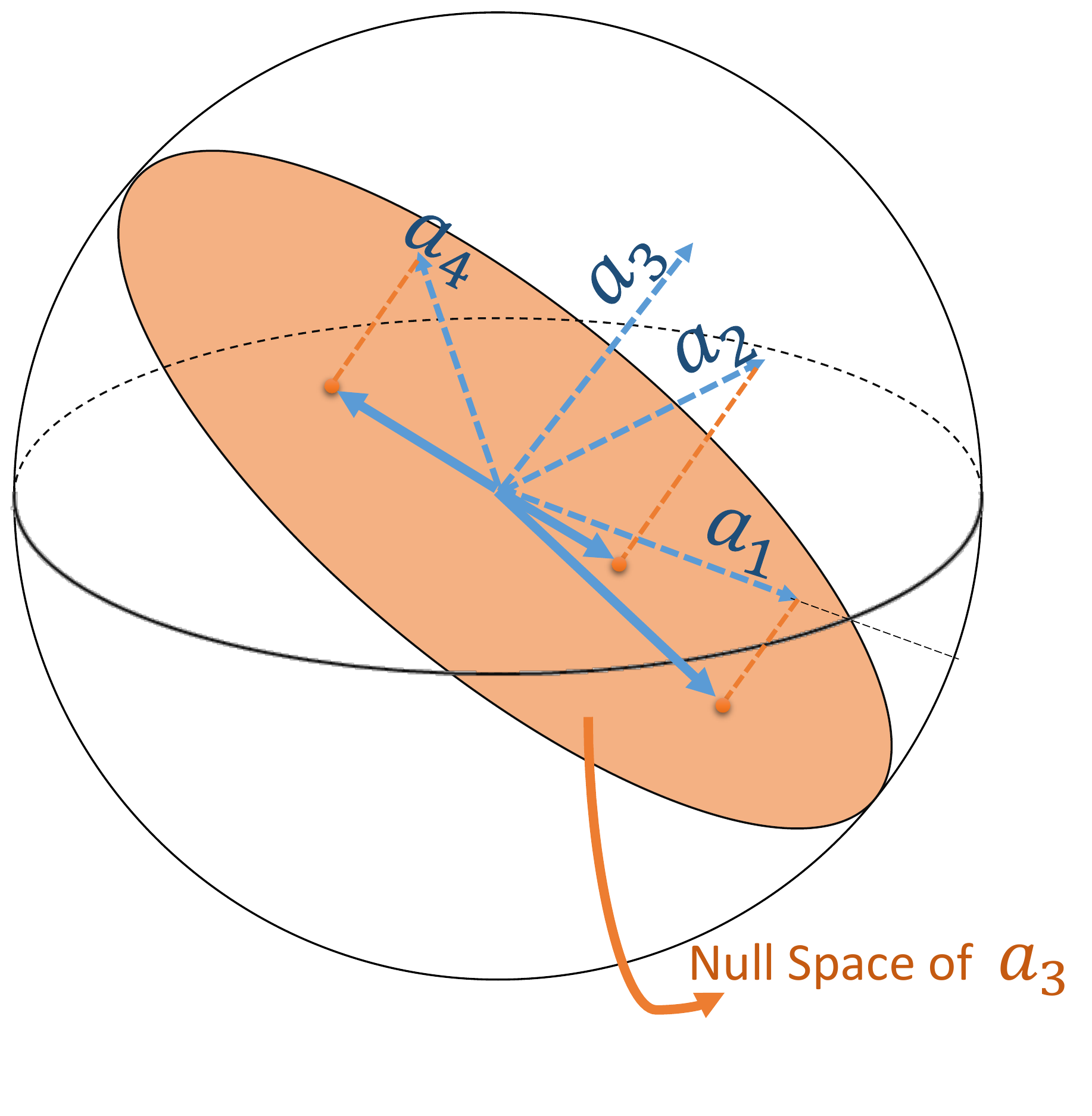}
\end{subfigure}
\caption{\small A toy example that illustrates the first iteration of IPM. (Left) The most matched sample with the first right singular vector, $\boldsymbol{v}$, is selected. (Right)  The rest of samples are projected on the null space of the selected sample in order to continue selection in the lower dimensional subspace.}
\label{fig:intution}
\end{figure} 


\subsection{A Lower Bound on Maximum Correlation}
In this section, we will derive a lower bound on the maximum  of  the  absolute  value  of  the  correlation  coefficient between data points $\boldsymbol{a}_1, \boldsymbol{a}_2,\ldots, \boldsymbol{a}_M$ and $\boldsymbol{v}$, when data are normalized on the unit sphere.  
Figure \ref{fig:intution}  shows an intuitive example for one iteration of the proposed algorithm. First, the leading singular vector is computed, and then the most correlated sample in the dataset is matched with the computed singular vector. Next, all data are projected onto the null space of the matched sample. The projected data are ready to perform one more iteration, if needed. 
These iterations are terminated either by reaching the desired number of selected samples or a given threshold of residual energy. Next, we present a lemma that guarantees the existence of a highly correlated sample with the first right singular vector. 

\begin{lem}
\label{lem:cor}
Let $\boldsymbol{a}_1, \boldsymbol{a}_2,\ldots, \boldsymbol{a}_M\in \mathbb{R}^N$ be $M$ given data points of dimension $N$. Let $\boldsymbol{A}$ denotes an $M\times N$ matrix with $\boldsymbol{a}_m^T$ being its $m^{th}$ row for $m=1,2,\ldots, M$. Let $\sigma_1$, $\boldsymbol{u}$ and $\boldsymbol{v}$ denote the first singular value, the corresponding left and right singular vectors of $\boldsymbol{A}$, respectively. Then, there exists at least one data point such that the absolute value of its inner product with $\boldsymbol{v}$ is greater than or equal to $\frac{\sigma_1}{\sqrt{M}}$. 
Hence,
\begin{equation}
\small
\underset{m}{\text{max}}\;| \boldsymbol{v}^T {\boldsymbol{a}}_m|\geq \frac{\sigma_1}{\sqrt{M}}.
\end{equation}
\end{lem}

The following proposition states a lower bound on the maximum of the absolute value of the correlation between data points $\boldsymbol{a}_1, \boldsymbol{a}_2,\ldots, \boldsymbol{a}_M$ and $\boldsymbol{v}$, when data are normalized on the unit sphere. First, let us define the following measure.

\begin{define}
\label{def:rom}
Rank-oneness measure (ROM) of a rank $R$ matrix $\boldsymbol{A}$ with singular values $\sigma_1, \sigma_2, \ldots, \sigma_R$ is defined as 

$$
\small
ROM(\boldsymbol{A})=\sqrt{\frac{\sigma_1^2}{\sum_{r=1}^R \sigma_r^2 }}=\frac{\sigma_1}{\|\boldsymbol{A}\|_F}.
$$
\end{define}
\begin{prop}
\label{pr:rom}
Assume the rows of $\boldsymbol{A}$ are normalized to lie on the unit sphere. There exists at least one data point, $i$, such that the correlation coefficient between $\boldsymbol{a_i}$ and the first right singular vector of $\boldsymbol{A}$ is greater than or equal to $\it{ROM}(\boldsymbol{A})$. 
\end{prop}

\subsection{Robustness to Perturbation}

Data selection algorithms are vulnerable to outlier samples. Since outlier samples are more spread in the space of data, their span covers a wider subspace. However, the spanned subspace by outliers may not be a proper representative subspace. DS3 adds a penalty to the cost function in order to reject outliers \cite{Elhamifar2016DissimilaritySelection}. Our proposed algorithm computes the first singular vector as the leading direction in each iteration. We  show here that this direction is the most robust spectral component against changes in the data. First consider the autocorrelation matrix  of data  defined as,
\small{
$$
\boldsymbol{C}=\sum_{m=1}^M \boldsymbol{a}_m \boldsymbol{a}_m^T.
$$
}
\normalsize{Eigenvectors of this matrix are equal to right sigular vectors of $\boldsymbol{A}$. Adding a new row in $\boldsymbol{A}$ does not change the size of matrix $\boldsymbol{C}$, but perturbs this matrix. The following lemma shows the robustness of  eigenvectors of $\boldsymbol{C}$ against perturbations. }

\begin{lem}
\label{lem:sens}
Assume square matrix $\boldsymbol{C}$ and its spectrum $[\lambda_i, v_i]$. Then, the following inequality holds,
\small{
$$
\|\partial v_i\|_2\le \sqrt{\sum_{j\neq i} \frac{1}{(\lambda_i-\lambda_j)^2}}\|\partial C\|_F.
$$
}
\end{lem}
\begin{define}
\normalsize{
The sensitivity coefficient of the $i^{\text{th}}$ eigenvector of a square matrix is defined by,}
\small{
$$
s_i\triangleq \sqrt{\sum_{j\neq i} \frac{1}{(\lambda_i-\lambda_j)^2}}.
$$
}
\end{define}

\normalsize{It is easy to show that $s_1<s_2$. Based on  Lemma \ref{lem:sens} and this definition the following proposition suggests a condition to satisfy $s_1<s_i,\;\forall i\ge2$. 
}

\begin{prop}
\label{pr:sens2}
Assume square matrix $\boldsymbol{C}$ and its spectrum $[\lambda_i, v_i]$, where  the gap between consecutive eigenvalues is decreasing. Then,
$$s_1<s_i, \;\;\forall i\ge 2.
$$
\end{prop}

 The proofs of Propositions and Lemmas in  this section are presented in the supplementary material. Moreover, the results of Proposition \ref{pr:rom}  and  \ref{pr:sens2} are also verified in supplementary material.

\begin{table*}
\small
\begin{center}
\begin{tabular}{|l|c|c|c|c|c|c|c|c|c|}
\hline
Average samples per class & 2 & 3 & 4 & 5 & 6 & 7 & 8 & 9 & 10\\
\hline\hline
Random   & 59.87 & 65.15 & 68.25 & 71.16 & 71.42 & 73.80 & 75.41 & 76.36 & 76.65 \\
Spectral Clustering    &  61.11 & 66.66 & 68.01 & 69.31 & 70.97 & 71.60 & 71.58 & 73.88 & 74.14\\
K-medoids & 62.51 & 66.29 &  69.36 &  69.28 & 71.55 & 73.06 & 73.96 & 75.25 & 75.20\\
DS3 \cite{Elhamifar2016DissimilaritySelection}    & 63.94 & 67.19 & 68.64 & 68.78 & 70.86 & 72.37 & 72.66 & 72.87 & 73.19\\
Uncertainty \cite{Gal2017DeepData} & 59.50 & 67.27 & 69.89 & 72.64 & 74.01 & 74.86 & 74.67 & 77.08 & 76.79 \\
IPM      & \textbf{64.93} & 68.86 & 71.02 & 72.74 & 73.71 & 73.93& 75.89 & 76.36 & 77.18  \\
IPM + Uncertainty   & \textbf{64.93} & 7\textbf{0.23} & \textbf{73.85} & \textbf{73.46} & \textbf{76.97} & \textbf{76.52} & \textbf{77.66} & \textbf{78.19} & \textbf{78.11} \\
\hline
\end{tabular}
\end{center}
\caption{\small Classification accuracy (\%) for action recognition on UCF-101 dataset, at different active learning cycles. The initial training set (cycle 1) is the same for all the methods. The accuracy for cycle 1 is $55.37\%$ and the accuracy using the full training set (9537 samples) is $82.23\%$.}
\label{tab:AL_results}
\end{table*}

\section{Applications of IPM}
\label{sec:experiments}
To validate our theoretical investigation and to empirically demonstrate the behavior and effectiveness of the proposed selection technique, we have performed extensive sets of experiments considering several different scenarios. We divide our experiments into three different subsections. 
In Section \ref{subsec:active-learning}, we use our algorithm in the \emph{active learning} setting and show that IPM is able to reduce the sampling cost significantly, by selecting the most informative unlabeled samples. 
Next, in Section \ref{subsec:reps}, we show the effectiveness of IPM in selecting the most informative representatives, by training the classifier using only a few representatives from each class.
Lastly, in Section \ref{subsec:exp_vid_sum}, the application of IPM for video summarization is exhibited. In addition, we investigate the robustness and other performance metrics, such as projection error and running time, of different selection methods and verify our theoretical results in the supplementary material.
\par 


\subsection{Active Learning}
\label{subsec:active-learning}

 \emph{Active learning} aims at addressing the costly data labeling problem by iteratively training a model using a small number of labeled data, and then querying the labels of some selected data, using an acquisition function. \par 

In active learning, the model is initially trained using a small set of labeled data (the initial training set). Then, the acquisition function selects a few points from the pool of unlabeled data, asks an oracle (often a human expert) for the labels, and adds them to the training set. Next, a new model is trained on the updated training set. By repeating these steps, we can collect the \emph{most informative} samples, which often result in significant reductions in the labeling cost. Now, the fundamental question in active learning is: Given a fixed labeling budget, what are the best unlabeled data instances to be selected for labeling for the best performance? \par 

In many active learning frameworks, new data points are selected based on the model uncertainty. However, the effect of such selection only kicks in after the size of the training set is large enough, so we can have a reliable uncertainty measure. In this section, we show that the proposed selection method can effectively find the best representatives of the data and outperforms several recent uncertainty-based and algebraic selection methods. \par 

In particular, we study IPM for active learning of video action recognition, using the 3D ResNet18 architecture\footnote{We use the code provided by the authors at \url{https://github.com/kenshohara/3D-ResNets-PyTorch}}, as described in \cite{Hara2017CanImageNet}. The experiments are run on UCF-101 human action dataset \cite{Soomro2012UCF101:Wild}, and the network is pretrained on Kinetics-400 dataset \cite{Kay2017TheDataset}. Here, we provide the results on split 1.

 To ensure that at least one sample per class exists in the training set, for the initial training, one sample per class is selected randomly and the fully-connected layer of the classifier is fine tuned. Then, at each active learning cycle, 
 one sample per class is selected, without the knowledge of the labels, and added to the training set. Next, using the updated training set, the fully connected layer of the network is fine tuned for $60$ epochs, using learning rate of $10^{-1}$, weight decay of $10^{-3}$, and batch size of $24$ on 2 GPUs. Rest of the implementation and training settings are the same as \cite{Hara2017CanImageNet}. Note that, in this experiment, fine-tuning is only performed to train the fully connected layer, because it achieved the best accuracy during the preliminary investigation for very small training sets, which is the scope of this experiment. Using this setting the classification accuracy using the full training set is $82.23\%$ \footnote{ State-of-the-art accuracy is $98\%$, which is achieved by full fine-tuning a two-stream I3D model\cite{Carreira2017QuoDataset}.}.
  \par 

The selection is performed on the convolutional features extracted from the last convolutional layer of the network. Table \ref{tab:AL_results} shows the accuracy of the trained network at each active learning cycle for different selection methods. In the table, DS3 stands for dissimilarity-based sparse subset selection, which has been proposed for finding an informative subset of a collection of data points\footnote{We use the code provided by the authors at \url{http://www.ccs.neu.edu/home/eelhami/codes.htm}} \cite{Elhamifar2016DissimilaritySelection}. The high computational complexity of DS3 prevents its implementation on all the data. So, we provide the results for DS3 only for the clustered version, meaning that one sample per cluster is selected using DS3. For spectral clustering results, the extracted features are clustered into 101 clusters, and one sample from each cluster is selected randomly. Furthermore, for uncertainty-based selection, Bayesian active learning, as described in \cite{Gal2017DeepData,Gal2016DropoutLearning}, is utilized. For that, a dropout unit with parameter $0.2$ is added before the fully-connected layer and the uncertainty measure is computed by using $10$ forward iterations (following the implementation in \cite{Gal2016DropoutLearning}). In our experiments, we use variation ratio\footnote{Variation ratio of $x$ is defined as $ 1 - \max_y p(y|x)$. which measures lack of confidence.} as the uncertainty metric, which is shown to be the most reliable metric among several well-known metrics \cite{Gal2017DeepData}. Also, for a fair comparison, the initial training set is the same for all the experiments.\par 

It is evident that, during the first few cycles, since the classifier is not able to generate reliable uncertainty score, uncertainty-based selection does not lead to a performance gain. In fact, random selection outperforms uncertainty-based selection. \par 

On the other hand,  IPM is able to select the critical samples. In the first few active learning cycles, IPM is constantly outperforming other methods, which translates into significant reductions in labeling cost for applications such as video action recognition. \par 
As the classifier is trained with more data, it is able to provide us with better uncertainty scores. Thus to enjoy the benefits of both IPM and uncertainty-based selection, we can use a compound selection criterion. For the extremely small datasets, samples should be selected only using IPM. However, as we collect more data, the uncertainty score should be integrated into the decision making process. Our proposed selection algorithm, unlike other methods, easily lends itself to such modification. At each selection iteration, instead of selecting the most correlated data with $\boldsymbol{v}$ (line 3 in Algorithm \ref{alg:IPM}), we can select the samples based on the following criterion:  
$$
m^*=\underset{m}{\arg \max} ~ \alpha \; |\boldsymbol{v}^T \Tilde{\boldsymbol{a}}_m|  + (1 - \alpha) \; q(\boldsymbol{a}_m),
$$
where $q(.)$ is an uncertainty measure, e.g. variation ratios. Parameter $\alpha$ determines the relative importance of the IPM metric versus the uncertainty metric. To gradually increase the impact of $q(.)$, as the model becomes more reliable, we start by setting $\alpha = 1$ and multiply it by decay rate of $0.95$ at each active learning cycle. As it is evident in Table \ref{tab:AL_results}, this compound selection criteria leads to better results for larger dataset sizes. 


\subsection{Learning Using Representatives}\label{subsec:reps}
In this experiment, we consider the problem of learning using representatives. We find the best representatives for each class and use this reduced training set for learning. Finding representatives reduces the computation and storage requirements, and can even be used for tasks such as clustering. In the ideal case, if we collect the samples that contain enough information about the distribution of the whole data set, the learning performance would be very close to the performance using all the data.

\subsubsection{Finding Representatives for Multi-PIE Dataset}\label{subsec:reps_multipie}

\begin{figure}
\centering
\begin{subfigure}{1\columnwidth}
\centering
\includegraphics[width=0.8\columnwidth]{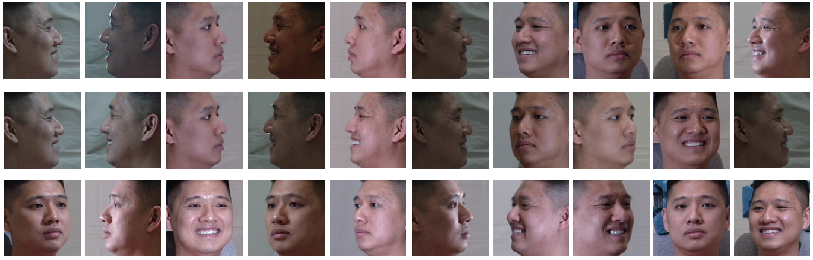}
   \caption{ \small The first row is obtained by K-medoids and the second and the third row show the selection of DS3 and IPM, respectively.}
\label{subfig:PIE-example}
\end{subfigure}
\begin{subfigure}[b]{1\columnwidth}
\centering
\includegraphics[width=0.6\columnwidth]{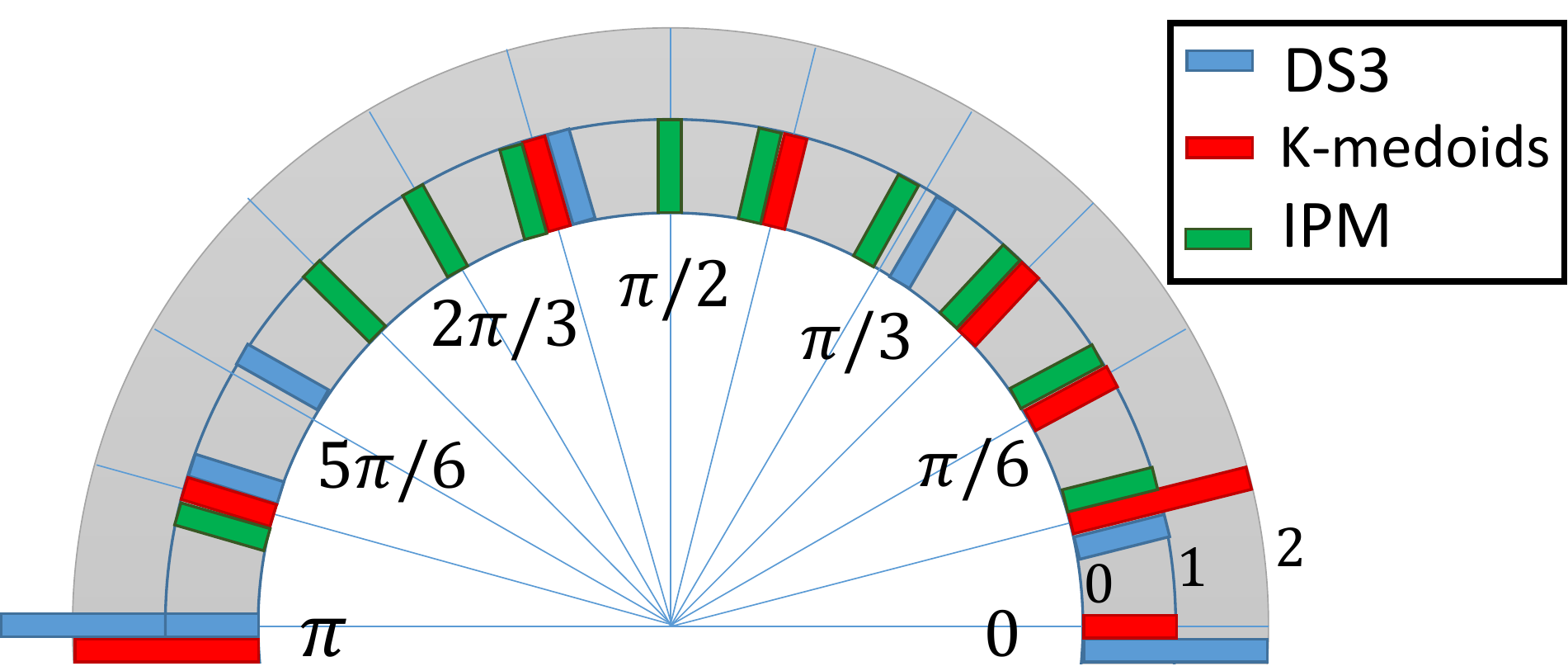}
   \caption{ \small Angles of the selected images. K-medoids selects 8 different angles. DS3 algorithm selects from 7  angles and our proposed IPM selects the maximum possible 10 distinguished angles.}
\label{subfig:angles}
\end{subfigure}
\caption{ \small Selection of 10 representatives out of 520 images of a subject and their corresponding angles. }
\label{fig:PIE_selection}
\end{figure}

\begin{figure}
\begin{subfigure}[b]{0.48\columnwidth}
   \includegraphics[width=1\columnwidth]{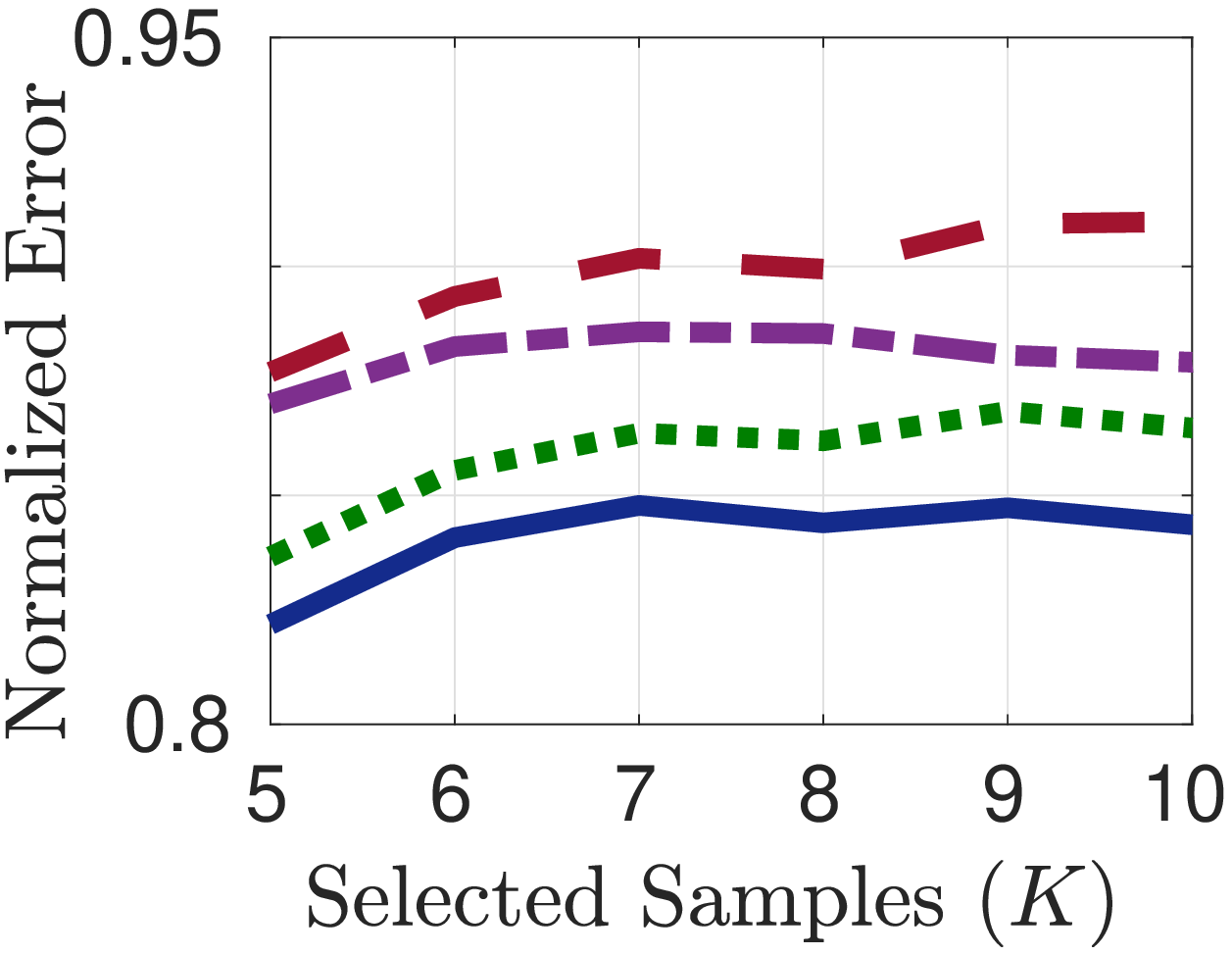}
\end{subfigure}
\begin{subfigure}[b]{0.48\columnwidth}
    \centering     
   \includegraphics[width=1\columnwidth]{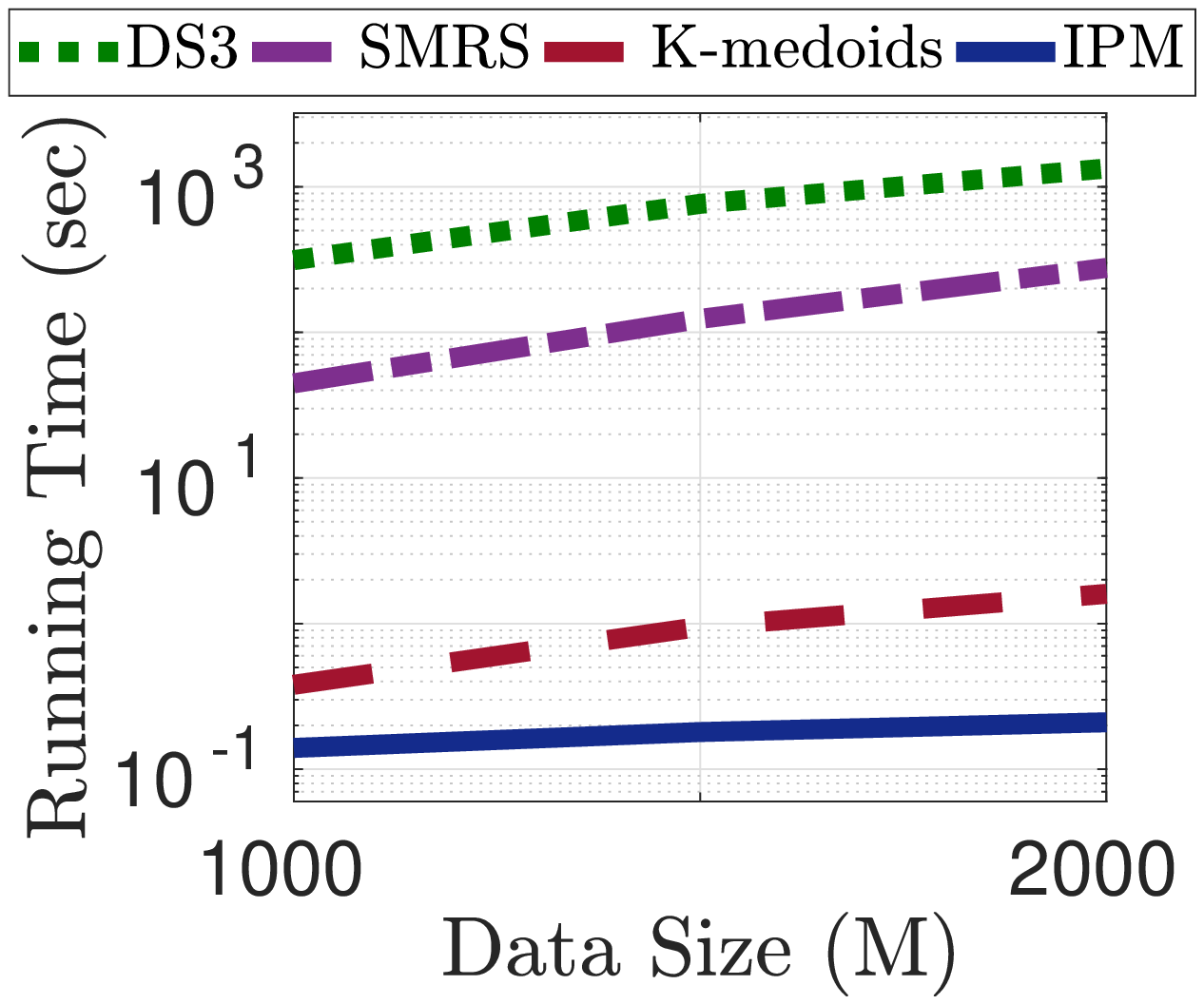}
\end{subfigure}
\caption{\small Performance of different methods for minimizing the main cost function of representative selection in equation  (\ref{eq:orig_sel}). (Left) The ratio of projection error using selection algorithms to  projection error of random selection for selecting $K$ representatives from each subject, averaged over all the subjects. (Right)  Running time of different algorithms versus number of input samples for selection.}
\label{fig:time_n_error}
\end{figure}

Here, we present our experimental results on CMU Multi-PIE Face Database \cite{Gross2010Multi-PIE}. We use 249 subjects from the first session with 13 poses, 20 illuminations, and two expressions. Thus, there are $13\times 20\times 2$ images per subject. Figure \ref{subfig:PIE-example} shows 10 selected images from 520 images of a subject. As it can be seen, the results of K-medoids and DS3 algorithms are concentrated on side views, while our selection provides images from more diverse angles. Figure\ref{subfig:angles} highlights this by showing  the angles of selected images of each algorithm. IPM selects from 10 different angles, while the selected images by DS3 and K-medoids contain repetitious angles.
Figure \ref{fig:time_n_error} shows the performance of different selection algorithms in terms of normalized projection error and running time. It is evident that our proposed approach  finds a better minimizer for Problem defined in equation (\ref{eq:orig_sel})  and is able to do so in orders of magnitude less time.

\subsubsection{Representatives To Generate Multi-view Images Using GAN}
\label{sec:reps_multipie_gan}
Next, to investigate the effectiveness of the proposed selection, we use the selected samples to train a generative adversarial network (GAN) to generate multi-view images from a single-view input. For that, the GAN architecture proposed in \cite{Tian2018CR-GAN:Generation} is employed. Following the experiment setup in \cite{Tian2018CR-GAN:Generation}, only $9$ poses between $\frac{\pi}{6}$ and $\frac{5\pi}{6}$ are considered. Furthermore, the first $200$ subjects are for
training and the rest are for testing. Thus, the total size of the training set is $72,000$, $360$ per subject. All the implementation details are same as \cite{Tian2018CR-GAN:Generation}, unless otherwise is stated\footnote{We use the code provided by the authors at \url{https://github.com/bluer555/CR-GAN}}.\par 

\begin{figure}
\includegraphics[width=1\columnwidth]{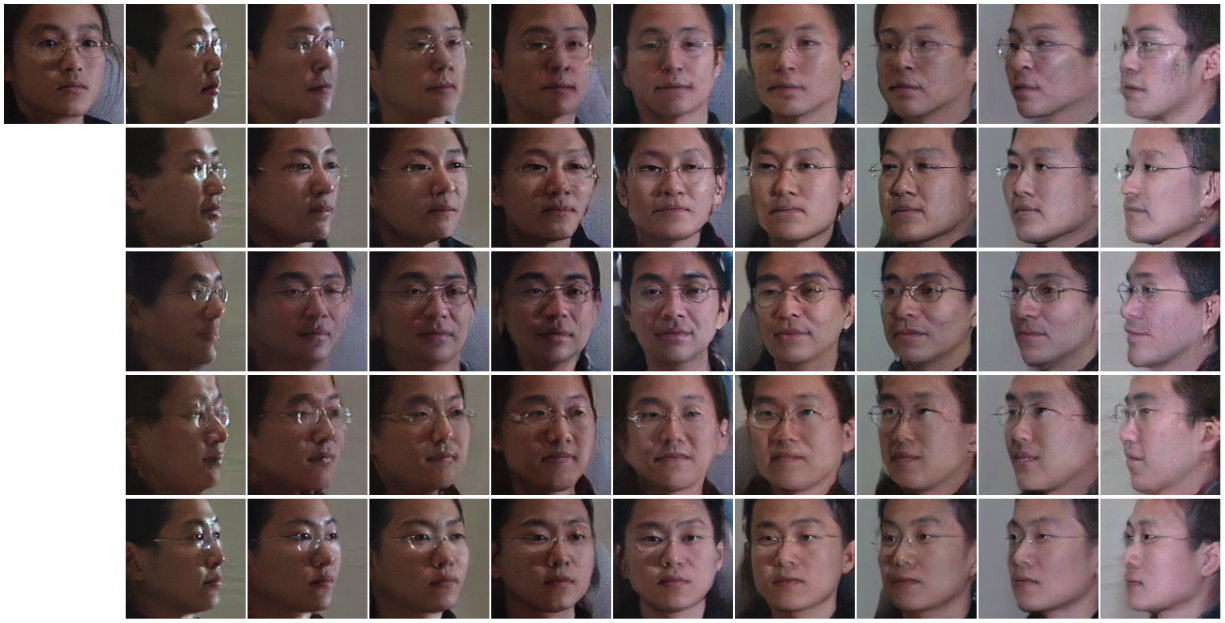}
\caption{\small Multi-view face generation results for a sample subject in Multi-PIE \cite{Gross2010Multi-PIE} testing set using CR-GAN \cite{Tian2018CR-GAN:Generation}. The network is trained on reduced training set ($9$ images per subject) using random selection (first row), K-medoids (second row), DS3 \cite{Elhamifar2016DissimilaritySelection} (third row), and IPM (fourth row). The fifth row shows the results generated by the network trained on all the data ($360$ images per subject). IPM-reduced dataset generates closest results to the complete dataset.}
\label{fig:PIE-GAN}
\end{figure}

We select only $9$ images from each subject ($1800$ total), and train the network with the reduced dataset for $300$ epochs using the batch size of $36$. Figure \ref{fig:PIE-GAN} shows the generated images of a subject in the testing set, using the trained network on the reduced dataset, as well as using the complete dataset. The network trained on samples selected by IPM (fourth row) is able to generate more realistic images, with fewer artifacts, compared to other selection methods (rows 1-3). Furthermore, compared to the results using all the data (row 5), it is clear that IPM-reduced dataset generates the closest results to the complete dataset. This is because, as demonstrated in Figure \ref{fig:PIE_selection}, samples selected by IPM cover more angles of the subject, leading better training of the GAN. See supplementary material for further experiments and sample outputs. \par 

\begin{table}
\begin{center}
\footnotesize
\begin{tabular}{|l|c|c|c|c|}
\hline
Method & Random & K-Medoids & DS3 & IPM  \\
\hline
$9$ images / subject & 0.5616 & 0.5993 & 0.6022 & {\bf 0.553}\\
\hline
$360$ images / subject & \multicolumn{4}{|c|}{0.5364}\\
\hline
\end{tabular}
\end{center}
\caption{ \small Identity dissimilarities between real and generated images by network trained on reduced (using different selection methods) and complete dataset.
}
\label{tab:GAN}
\end{table}


For a quantitative performance investigation, we evaluate the identity similarities between the real and generated images. For that, we feed each pair of real and generated images to a ResNet18\footnote{We use the naive ResNet18 architecture as described in \cite{Cao2018Pose-RobustMapping}.}, trained on MS-Celeb-1M dataset \cite{Guo2016MS-Celeb-1M:World}, and obtain 256-dimensional features. $\ell_2$ distances of features correspond to the face dissimilarity. Table \ref{tab:GAN} shows the normalized $\ell_2$ distances between the real and generated images, averaged over all the images in the testing set. Our method outperforms other selection methods in this metric as well. Thus, from Figure \ref{fig:PIE-GAN} (qualitative) and Table \ref{tab:GAN} (quantitative), we can conclude that the IPM-reduced training set contains more information about the complete set, compared to other selection methods. \par 


\begin{table}
\small
\begin{center}
\begin{tabular}{|l|c|c|c|c|c|c|c|}
\hline
\footnotesize{Samples / Class} & 1 & 2 & 3 & 4 & 5 & 6   \\
\hline\hline
Random   & 54.6 & 64.7 & 69.2 & 70.5 & 72.9 & 74.0  \\
K-medoids & 61.0  & 67.7 & 69.4 & 70.9 & 71.7 & 72.0  \\
DS3\cite{Elhamifar2016DissimilaritySelection}     & 60.8 & 69.1 & 74.0 & 75.2 & 74.8 & 75.3   \\
 IPM      & \textbf{65.3} & \textbf{72.6} & \textbf{74.9} & \textbf{77.6} & \textbf{77.0} & \textbf{78.5}  \\
\hline
\end{tabular}
\end{center}
\caption{ \small  Accuracy (\%) of ResNet18 on UCF-101 dataset, trained using only the representatives selected by different methods. The accuracy using the full training set (9537 samples) is $82.23\%$.}
\label{tab:reps}
\end{table}

\subsubsection{Finding Representatives for UCF-101 Dataset}
\label{subsec:reps_ucf101}
Here, similar to Section \ref{subsec:active-learning}, we use a 3D ResNet18 classifier pretrained on Kinetics-400 dataset, and the selection algorithms are performed on feature space generated by the output of the last convolutional layer. 
 To find the representatives, we use the selection methods to sequentially find the most informative representatives from each class. After selecting the representatives, the fully connected layer of the network is finetuned in the same manner as described in Section \ref{subsec:active-learning}. Table \ref{tab:reps} shows the performance of different selection methods for different numbers of representatives per class. As more samples are collected, the performance gap among different methods, including random, decreases. This is expected, since finding only one representative for each class is a much more difficult task, compared to choosing many, e.g. $6$, representatives. \par 

Using only one representative selected by IPM, we can achieve a classification accuracy of 65.3\%, which is more than 10\% improvement compared to random selection and more than 4\% improvement compared to other competitors. 
\par 


\begin{figure}
\centering     
\begin{subfigure}[c]{0.25\columnwidth}
    \centering     
   \includegraphics[width=1\columnwidth]{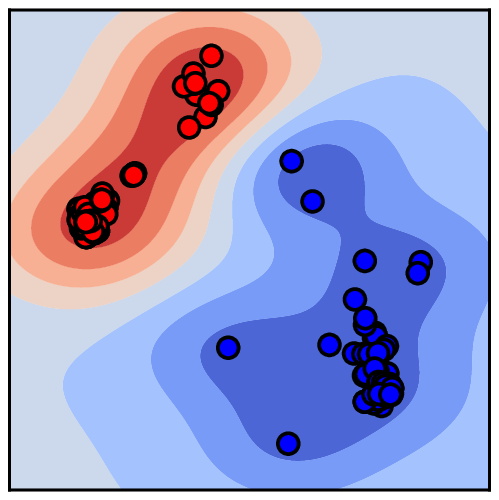}
   \caption{\small}
   \label{subfig:tsne-orig}
\end{subfigure}
\begin{subfigure}[c]{0.6\columnwidth}
    \centering     
   \includegraphics[width=1\columnwidth]{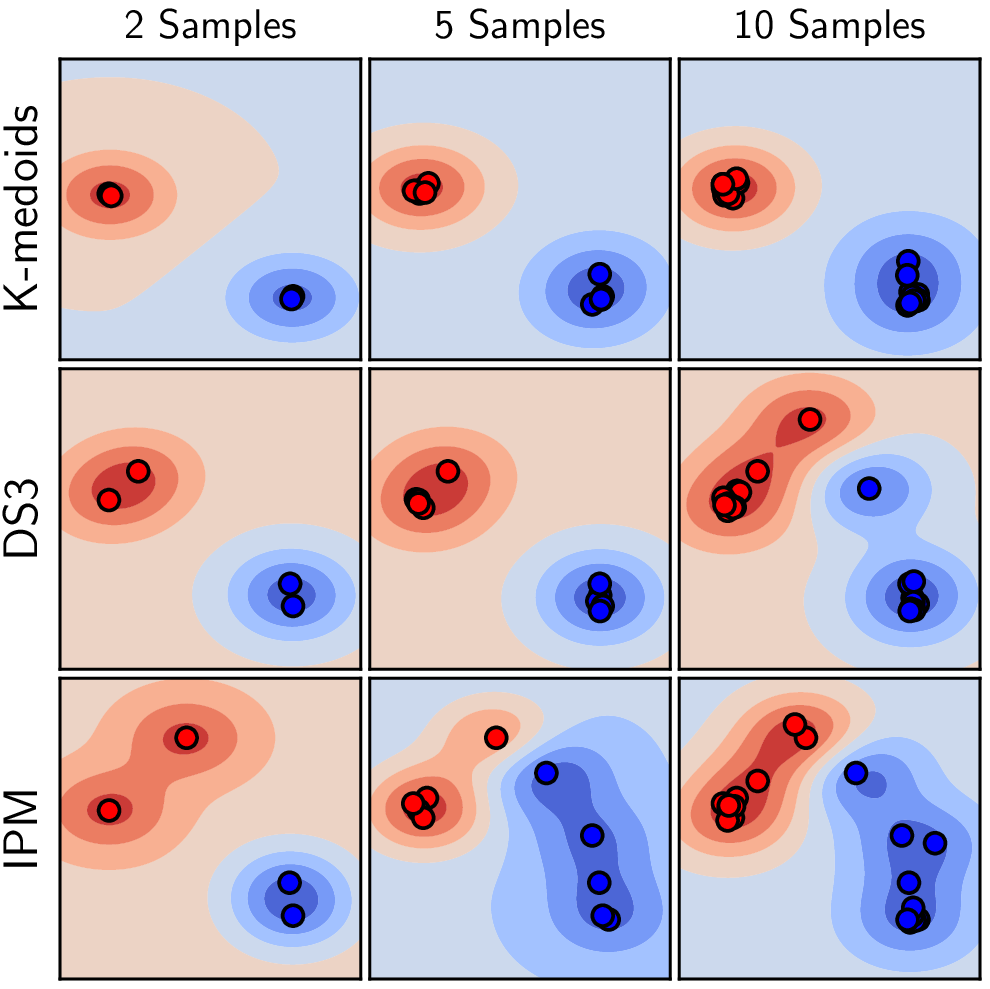}
   \caption{\small }
   \label{subfig:tsne-selections}
\end{subfigure}
\caption{\small t-SNE visualization  \cite{LaurensvanderMaaten2014VisualizingT-SNE} of  two randomly selected classes of UCF-101 dataset and their representatives selected by different methods. (\subref{subfig:tsne-orig}) Decision function learned by using all the data. The goal of selection is to preserve the structure with only a few representatives. (\subref{subfig:tsne-selections}) Decision function learned by using 2 (first column), 5 (second column), and 10 (third column) representatives per class, using K-medoids (first row), DS3 \cite{Elhamifar2016DissimilaritySelection} (second row), and IPM (third row). IPM can capture the structure of the data better using the same number of selected samples.}
\label{fig:tsne_ucf101}
\end{figure} 

Figure \ref{fig:tsne_ucf101} shows the t-SNE visualization \cite{LaurensvanderMaaten2014VisualizingT-SNE} of the selection process for two randomly selected classes of UCF-101. 
Selection is performed on the original $512$-dimensional feature space. 
To visualize the structure of the data, the contours represent the decision function of an SVM trained in this 2D space. This experiment illustrates  that each IPM sample contains new structural information, as the selected samples are far away from each other in the t-SNE space, compared to other  methods. Moreover, it is evident that as we collect more samples, the structure of the data is better captured by the samples selected by IPM, compared to other methods selecting the same number of representatives. The decision boundaries of the classifier trained on $5$ IPM-selected samples look very similar to the  boundaries learned from all the data. This leads to significant accuracy improvements, as already discussed and exhibited in Table \ref{tab:reps}.

\begin{table}
\begin{center}
\small
\begin{tabular}{|l|c|c|c|c|}
\hline
Images per Class& 1 & 5 & 10 & 50  \\
 & (0.08\%)& (0.4\%)&(0.8\%) &(4\%) \\
\hline
\hline
Random & 3.18 & 8.71 & 12.97 & 25.61   \\
\hline
K-Medoids & 11.78 & 17.01 & 17.56 & 26.86 \\
\hline
IPM & 12.50 & 21.69 & 25.26 & 30.77  \\
\hline 
\end{tabular}
\end{center}
\caption{ \small Top-1 classification accuracy (\%) on  ImageNet, using selected representatives from each class. Accuracy using all the labeled data (~1.2M samples) is $46.86\%$. Numbers in () show the size of the selected representatives as a \% of the full training set.
}
\label{tab:ImageNet}
\end{table}

\subsubsection{Finding Representatives for ImageNet}
\label{subsec:reps_Imagenet}

In this section, we use ImageNet dataset \cite{Deng2009ImageNet:Database}
to show the effectiveness of IPM in selecting the representatives for image classification task. For that, first, we extract features from images in an unsupervised manner, using the method proposed in \cite{Wu2018UnsupervisedDiscrimination}. We then we perform selection in the learned $128$-dimensional space and perform $k$-nearest neighbors ($k$-NN) using the learned similarity metric, following the experiments in \cite{Wu2018UnsupervisedDiscrimination} \footnote{We use the feature space generated by the ResNet50 backbone, as provided in \url{https://github.com/zhirongw/lemniscate.pytorch}}. Here, we show that we can learn the feature space and the similarity metric in an unsupervised manner, as there is no shortage of unlabeled data, and use only a few labeled representatives to classify the data. \par 

Due to the volume of this dataset, selection methods based on convex-relaxation, such as DS3 \cite{Elhamifar2016DissimilaritySelection} and SMRS \cite{Elhamifar2012SeeObjects}, fail to select class representatives in a tractable time (as discussed before and shown in Figure \ref{fig:time_n_error} for Multi-PIE dataset). Table \ref{tab:ImageNet} shows the top-1 classification accuracy for the testing set using $k$-NN. Using less than $1\%$ of the labels, we can achieve an accuracy of more than 25\%, showing the potential benefits of the proposed approach for dataset reduction.   Classification accuracy of $k$-NN, using the learned similarity metric, reflects the representativeness of the selected samples, thus highlighting the fact that IPM-selected samples preserve the structure of the data fairly well. \par 

\begin{table}
\begin{center}
\small
\begin{tabular}{|l|c|c|}
\hline
Method & F-measure & Recall  \\
\hline
\hline
\multicolumn{3}{|l|}{Selection Methods (Unsupervised)} \\
\hline
Random & 26.30 & 23.73\\
Uniform & 28.68 & 25.76\\
\hline
K-medoids & 30.11 & 27.30 \\
DS3 & 30.13 & 27.34\\
{\bf IPM} & {\bf 31.53} & {\bf 29.09} \\
\hline
\hline
\multicolumn{3}{|l|}{Supervised Summarization Methods} \\
\hline
SeqDPP \cite{Gong2014DiverseSummarization} & 28.87 & 26.83 \\
Submod-V \cite{Gygli2015VideoObjectives} & 29.35 & 27.43 \\
Submod-V+ \cite{Plummer2017EnhancingEmbedding} & 34.15 & 31.59 \\

\hline
\end{tabular}
\end{center}
\vspace{-2mm}
\caption{ \small F-measure and recall scores using ROUGE-SU metric for UT Egocentric video summarization task.  Results are reported for several supervised and unsupervised methods. 
}
\vspace{-2mm}
\label{tab:vid_sum}
\end{table}
\vspace{-3mm}
\subsection{Video Summarization}
\label{subsec:exp_vid_sum}

In this section, we evaluate the performance of the proposed selection algorithm on the video summarization task. The goal is to select key frames/clips and create a video summary, such that it contains the most essential contents of the video. We evaluate our approach on UT Egocentric (UTE) dataset \cite{YongJaeLee2012DiscoveringSummarization,Lu2013Story-DrivenVideo}. It contains 4 first-person videos of 3-5 hours of daily activities, recorded in an uncontrolled environment. Authors in \cite{Yeung2014VideoSET:Text} have provided text annotations for each 5-second segment of the video, as well as human-provided reference summaries for each video. Following \cite{Plummer2017EnhancingEmbedding,Gygli2015VideoObjectives,Yeung2014VideoSET:Text}, the performance is evaluated in text domain. For that, a text summary is created by concatenating the text annotations associated with the selected clips. The generated summaries are compared with the reference summaries using the ROUGE
metric \cite{Lin2004Rouge:Summaries}. As in prior work, we report f-measure and recall using the ROUGE-SU score with the same parameters as in \cite{Plummer2017EnhancingEmbedding,Gygli2015VideoObjectives,Yeung2014VideoSET:Text}. \par 

Table \ref{tab:vid_sum} provides the results for two-minute-long summaries (24 5-second samples), generated by different methods. To generate results using K-medoids, DS3, and IPM, we use $1024$-dimensional feature vectors extracted using GoogleNet \cite{Szegedy2015GoingConvolutions}, as described in \cite{Zhang2016VideoMemory}. Then, the features are clustered into 24 clusters using K-means and one sample is selected from each cluster using different selection techniques. The results  are the mean results over all the $4$ videos and over $100$ runs. Furthermore, for the supervised methods, the results are as reported in \cite{Plummer2017EnhancingEmbedding}. {\em The proposed unsupervised selection method, IPM, is the closest competitor to the state-of-art supervised method proposed in \cite{Plummer2017EnhancingEmbedding}, outperforming other unsupervised methods and some of the supervised methods}. These supervised methods split the dataset into training, and testing sets and use reference text or video summaries of the training set to learn to summarize the videos from the test set. This  experiment demonstrates the strength of IPM and the potential benefits of employing it in more advanced unsupervised or supervised schemes.  \par 

\section{Conclusions}
\label{sec:conclusion}

 A novel data selection algorithm, referred to as Iterative Projection and Matching (IPM) is presented, that selects the most informative data points in an iterative and greedy manner. 
Interestingly, we show that our greedy approach, with linear complexity wrt the dataset size, is able to outperform state-of-the-art methods, which are based on convex relaxation, in several performance metrics such as projection error and running time.
Furthermore, the effectiveness and compatibility of our approach are demonstrated in a wide array of applications such as active learning, video summarization, 
and learning from representatives. This motivates us to further investigate the potential benefits and applications of IPM in other computer vision problems. \par 

\section{Acknowledgements}
This research is based upon work supported in parts by the National Science Foundation under Grants No. 1741431 and CCF-1718195 and the Office of the Director of National Intelligence (ODNI), Intelligence Advanced Research Projects Activity (IARPA), via IARPA R\&D Contract No. D17PC00345. The views, findings, opinions, and conclusions or recommendations contained herein are those of the authors and should not be interpreted as necessarily representing the official policies or endorsements, either expressed or implied, of the NSF, ODNI, IARPA, or the U.S. Government. The U.S. Government is authorized to reproduce and distribute reprints for Governmental purposes notwithstanding any copyright annotation thereon.

{\small
\bibliographystyle{plain}
\bibliography{mendeley.bib}
}
\clearpage
\title{Supplementary Material}
\author{}
\maketitle
\setcounter{section}{0}
The supplementary material provides the proofs of the stated theoretical results and more experimental results are referred to another document mentioned in the footnote.

\section{Proofs}
\label{sec:supp_proofs}

\noindent{\bf Proof of Lemma \ref{lem:cor}}: 
The inner product between data points of $\boldsymbol{a}_m$'s and $\boldsymbol{v}$ can be calculated as the elements of $\boldsymbol{A}\boldsymbol{v}$. Taking the SVD of $\boldsymbol{A}$, we have $A=\boldsymbol{U\Sigma V}^T$ and
$$
|\boldsymbol{A}\boldsymbol{v}|= |\boldsymbol{U\Sigma V}^T\boldsymbol{v}|=\sigma_1 |\boldsymbol{u}|.
$$
Here, $|.|$ means the element-wise absolute value operation. Since $\|\boldsymbol{u}\|_2=1$, there exist at least one element in $\boldsymbol{u}$ (denote as $u_i$) for which $|u_i|\geq \frac{1}{\sqrt{M}}$. Therefore,
\begin{equation}\label{inner}
\small
|\boldsymbol{v}^T \boldsymbol{a}_i|\ge \frac{\sigma_1}{\sqrt{M}}.
\end{equation}
This clearly leads to 
\begin{equation}
\small
\underset{m}{\text{max}}\;| \boldsymbol{v}^T {\boldsymbol{a}}_m|\geq \frac{\sigma_1}{\sqrt{M}}.
\end{equation}
$$
$$

\noindent{\bf Proof of Proposition \ref{pr:rom}}:

According to Lemma~\ref{lem:cor}, there exists a data point $i$ for which (\ref{inner}) holds. Given that $\boldsymbol{v}$ has unit length and the rows of $\boldsymbol{A}$ are normalized we have
\begin{equation*}
    |\rho_i|=\frac{|\boldsymbol{v}^T \boldsymbol{a}_i|}{||\boldsymbol{v}||_2||\boldsymbol{a}_i||_2}\geq\frac{\sigma_1}{\sqrt{M}}=\frac{\sigma_1}{\|\boldsymbol{A}\|_F}=ROM(\boldsymbol{A}),
\end{equation*}
where $\rho_i$ is the correlation between two vectors $\boldsymbol{v}$ and $\boldsymbol{a_i}$ and ${\|\boldsymbol{A}\|_F}={\sqrt{\sum_{j=1}^{M}||\boldsymbol{a}_j^T||_2^2}}={\sqrt{M}}$. Accordingly, 
\begin{equation}
\underset{m}{\text{max}}\;|\rho_m |\geq ROM(\boldsymbol{A}).
\end{equation}

\noindent{\bf Proof of Lemma \ref{lem:sens}}:

First we compute the derivative of $i^{\text{th}}$ eigenvector in terms of  matrix $\boldsymbol{C}$ \cite{RUDISILL1974DerivativesMatrix},
\begin{equation}
\small
\label{eq:proof1}
\partial \boldsymbol{v}_i =(\sigma_i^2 I - C)^+ \partial \boldsymbol{C} \boldsymbol{v}_i,
\end{equation}

\noindent where $(.)^+$ indicates the Moore–Penrose inverse operator. Matrix $\sigma_i^2 I - C$ is singular and its Moore–Penrose inverse can be written as follows,
$$
\small
(\sigma_i^2 I - C)^+=\boldsymbol{V}\boldsymbol{\Sigma}_i \boldsymbol{V}^T,
$$
where, diagonal elements of $\boldsymbol{\Sigma}_i$ is equal to $1/(\boldsymbol{\lambda}-\lambda_i)$ except the $i^{\text{th}}$ diagonal element which is equal to $0$. Vector $\boldsymbol{\lambda}$ includes the eigenvalues of $\boldsymbol{C}$. Taking $\ell_2$ norm from both side of (\ref{eq:proof1}) we have
\begin{align}
\small
\|\partial \boldsymbol{v}_i\|_2 &=\|\boldsymbol{V} \boldsymbol{\Sigma}_i \boldsymbol{V}^T \partial \boldsymbol{C} \boldsymbol{v}_i\|_2\nonumber\\
&\le \| \Sigma_i\|_F\; \|\partial \boldsymbol{C}\|_F=\sqrt{\sum_{j\neq i} \frac{1}{(\lambda_i-\lambda_j)^2}}\|\partial C\|_F.\nonumber
\end{align}
Note that $\boldsymbol{V}$ is unitary and $\boldsymbol{v}_i$ is normalized and $\sigma_i^2=\lambda_i$.

\textbf{Proof of Proposition \ref{pr:sens2}}:
Obviously $\lambda_1>\lambda_2$ implies that $s_1<s_2$. Let us write the expansion of $s_1$ and $s_i\;$ for $i>2$.
\vspace{-2mm}
\begin{align}
\small
s_1=&\frac{1}{(\lambda_1-\lambda_2)^2}+\frac{1}{(\lambda_1-\lambda_3)^2}+\cdots\nonumber \\&\frac{1}{(\lambda_1-\lambda_i)^2}+\cdots+\frac{1}{(\lambda_1-\lambda_N)^2}\nonumber
\end{align}
\vspace{-2mm}
\begin{align}
\small
s_i=\frac{1}{(\lambda_i-\lambda_1)^2}+\frac{1}{(\lambda_i-\lambda_2)^2}+\cdots+\frac{1}{(\lambda_i-\lambda_N)^2}\nonumber
\end{align}
The $(i-1)^{\text{th}}$ term of $s_1$ is equal to the first term of $s_i$. As eigenvalues are sorted in descending order, $i^{\text{th}}$ to $N^{\text{th}}$ terms of $s_1$ are less than $i^{\text{th}}$ to $N^{\text{th}}$ terms of $s_i$, correspondingly. Thus it is sufficient to show that,
$$
\sum_{j=2}^{i-1} \frac{1}{(\lambda_1-\lambda_j)^2}<\sum_{j=2}^{i-1} \frac{1}{(\lambda_j-\lambda_i)^2}.
$$

Which is immediately concluded if the gap between consecutive eigenvalues is decreasing. 

Please refer to the footnote link in order to see validating the theoretical results and more experiments\footnote{\url{http://cwnlab.eecs.ucf.edu/?page_id=577}}. Moreover, the implementation of IPM in Matlab and Python is available online as well as implementation of some presented applications of IPM. 

\end{document}